\documentclass[conference]{IEEEtran}

\IEEEoverridecommandlockouts
\usepackage{graphicx}

\usepackage[T1]{fontenc}    
\usepackage[utf8]{inputenc}

\usepackage{amsmath, graphicx, setspace}

\usepackage[lofdepth,lotdepth]{subfig}
\usepackage[]{caption}
\usepackage{multirow}
\usepackage{array}
\usepackage[colorinlistoftodos]{todonotes}
\usepackage{tabu}
\usepackage{subfig}
\usepackage{xcolor}
\definecolor{darkspringgreen}{RGB}{9, 115, 69}
\usepackage{nicefrac}       
\usepackage{microtype}      
\usepackage{fancyhdr}       
\usepackage{float}
\usepackage[hidelinks]{hyperref}
\usepackage[numbers,sort&compress]{natbib}

\usepackage{todonotes} 

\begin{document}

\title{Algılanan Stres Testinin Makine Öğrenmesi ile Analiz Edilmesi}

\author{
    \IEEEauthorblockN{\large Toygar Tanyel}
    \IEEEauthorblockA{  Bilgisayar Mühendisliği,
                        Yıldız Teknik Üniversitesi \\
                        \texttt{tanyel.med.ai@gmail.com} }
}

\maketitle

\begin{ozet}
Bu çalışmanın amacı, 150 kişinin algıladığı stres seviyesini belirlemek ve Türkçeye uyarlanmış sorulara verilen yanıtları makine öğrenmesi ile analiz etmektir. Test, 14 sorudan oluşur ve her bir soru 0 ile 4 arasındaki bir ölçekte puanlanır. Bu da toplam puan aralığını 0-56 olarak belirler. Bu sorulardan 7 tanesi negatif bağlamda formüle edilmiş ve bu doğrultuda puanlanırken, kalan 7 tanesi pozitif bağlamda formüle edilmiş ve puanlama ters şekilde yapılmıştır. Test ayrıca algılanan özyeterlik ve stres/rahatsızlık algısı olmak üzere iki alt faktörü belirlemek amacıyla tasarlanmıştır. Bu araştırmanın temel hedefleri, yapay zeka tekniklerini kullanarak test sorularının eşit öneme sahip olmayabileceğini göstermek, hangi soruların yanıtlanırken toplumsal anlamda farklılık gösterdiğini makine öğrenimi ile ortaya koymak ve sonuç olarak psikolojik olarak gözlemlenen belirgin kalıpların varlığını göstermektir. Bu çalışma, testin makine öğrenmesiyle tekrarlanması ile mevcut psikoloji literatüründen farklı bir perspektif sunmaktadır. Ayrıca, algılanan stres testinin sonuçlarını yorumlamada kullanılan ölçeğin doğruluğunu sorgulamakta ve test sorularının önem sıralamasındaki farklılıkların göz ardı edilmemesi gerektiğini vurgulamaktadır. Bu çalışmanın bulguları, stresle başa çıkma stratejileri ve terapötik yaklaşımlar açısından yeni bir bakış açısı sunmaktadır.
Kaynak kod: \texttt{\href{https://github.com/toygarr/ppl-r-stressed}{https://github.com/toygarr/ppl-r-stressed}}
\end{ozet}
\begin{IEEEanahtar}
algılanan stres, makine öğrenmesi.
\end{IEEEanahtar}

\begin{abstract}
The aim of this study is to determine the perceived stress levels of 150 individuals and analyze the responses given to adapted questions in Turkish using machine learning. The test consists of 14 questions, each scored on a scale of 0 to 4, resulting in a total score range of 0-56. Out of these questions, 7 are formulated in a negative context and scored accordingly, while the remaining 7 are formulated in a positive context and scored in reverse. The test is also designed to identify two sub-factors: perceived self-efficacy and stress/discomfort perception. The main objectives of this research are to demonstrate that test questions may not have equal importance using artificial intelligence techniques, reveal which questions exhibit variations in the society using machine learning, and ultimately demonstrate the existence of distinct patterns observed psychologically. This study provides a different perspective from the existing psychology literature by repeating the test through machine learning. Additionally, it questions the accuracy of the scale used to interpret the results of the perceived stress test and emphasizes the importance of considering differences in the prioritization of test questions. The findings of this study offer new insights into coping strategies and therapeutic approaches in dealing with stress.
Source code: \texttt{\href{https://github.com/toygarr/ppl-r-stressed}{https://github.com/toygarr/ppl-r-stressed}}
\end{abstract}

\newpage
\section{G{\footnotesize İ}r{\footnotesize İ}ş}

Stres, günümüzün hızlı ve tempolu yaşam tarzıyla birlikte yaygın bir sorun haline gelmiştir ve bireylerin fiziksel ve psikolojik sağlığını etkileyebilmektedir \cite{5schneiderman2005stress}. Bu nedenle, stres düzeyini objektif bir şekilde değerlendirebilen etkili ölçüm araçlarına ihtiyaç duyulmaktadır. Bu çalışmanın amacı, Türkçeye uyarlanan algılanan stres testini kullanarak 150 kişinin stres düzeyini tespit etme konusunda testteki soruların etkisini makine öğrenmesi yöntemleriyle değerlendirmektir.

Stres düzeyini ölçmek için kullanılan algılanan stres testindeki 14 sorunun hem pozitif hem de negatif olarak yanıt tepkisini ölçmesi, bireylerin algıladıkları stres düzeyini çeşitli yönlerden değerlendirmeyi hedeflemektedir. 

Bu çalışmada, algılanan stres testinin yanı sıra, yetersiz özyeterlik algısı ve stres/rahatsızlık algısı olmak üzere iki alt faktörü de belirlemek amacıyla çoklu etiket yapısı tasarlanmıştır. Makine öğrenmesi için hazırlanan skorlama sisteminin yapısı ve içeriği, bireylerin stres düzeyini tespit etmede stres durumunun ve faktörlerin öğrenim sırasında bağımsız rol alacağı şekilde kurgulanmıştır.

Araştırmanın temel hedefleri, yapay zeka tekniklerinin kullanılmasıyla test sorularının eşit öneme sahip olmayabileceğini göstermek, makine öğrenimiyle hangi soruların yanıtlanırken toplumsal açıdan farklılık gösterdiğini ortaya koymak ve sonuç olarak psikolojik açıdan gözlemlenen belirgin kalıpların varlığını sorular üzerinden tartışmaya açmaktır. Bu çalışma, mevcut psikoloji literatüründen \cite{1asberg2008structural, 2baltacs1998standardization, 3orucu2009psychometric, 4eskin1993reliability, 6cohen1983global, 7schiffrin2010stressed, 8reis2010perceived} farklı bir yaklaşım kullanarak, stres düzeyini değerlendirmede etkili bulunan \cite{eskin2013algilanan} algılanan stres testinin etkinliğini ölçmeyi amaçlamaktadır.

Bölüm 2'de verinin nasıl toplandığı, işlendiği ve skor sistemi anlatılmıştır. Bölüm 3'te veri analizi yapılırken, Bölüm 4'te makine öğrenmesi ile ilgili gerekli detaylar verilmiştir. Bölüm 5'te ise sonuçların özeti ve limitasyonlar paylaşılmaktadır.

\section{Materyal \& Metodlar}

\subsection{Veri Toplama}
Veri toplama işlemi Google Forms üzerinden gerçekleştirilmiştir ve girilen yanıtlar anonim bir şekilde bir excel dosyasına kaydedilmiştir. Verinin kitlesi ağırlıklı olarak üniversite öğrencileri ve onların yakınlarıdır. Ayrıca test, Türkiye Cumhurbaşkanlığı seçimlerinden önceki hafta uygulanmıştır (4 Mayıs - 11 Mayıs 2023). Psikoloji literatüründe bu testin sonucu için ölçüm aracı olarak "Algılanan Stres Ölçeği" \cite{eskin2013algilanan} kullanılabilmektedir.

\begin{figure*}[]
\subfloat[\centering Ham veri\vspace{0.5cm}]{{\includegraphics[width=\textwidth]{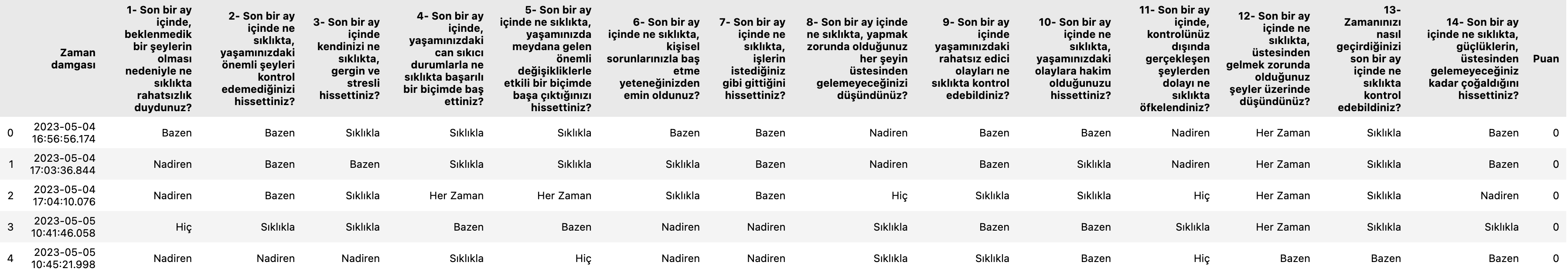} }}%
    \quad
    \subfloat[\centering Düzenlenmiş veri]{{\includegraphics[width=\textwidth]{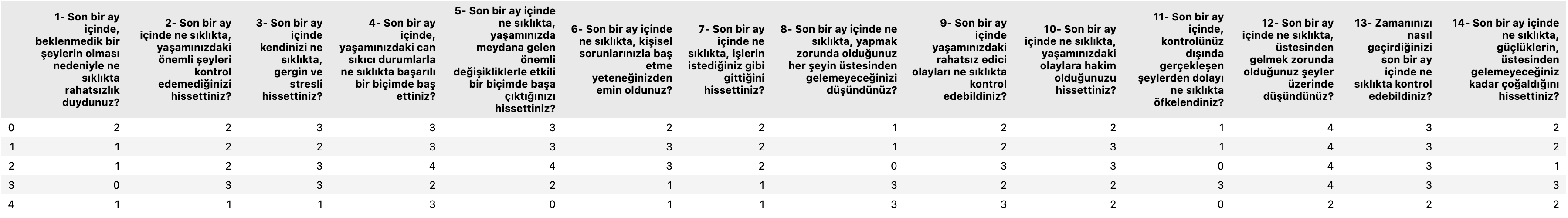} }}
\caption{Stres skorlarını hesaplamadan önce veri seti yapısı.} \label{fig:dataset}%
\end{figure*}

\todo[inline, color=gray!5]{\textbf{Sorular}\newline

1- Son bir ay içinde, beklenmedik bir şeylerin olması nedeniyle ne sıklıkta rahatsızlık duydunuz?

2- Son bir ay içinde ne sıklıkta, yaşamınızdaki önemli şeyleri kontrol edemediğinizi hissettiniz?

3- Son bir ay içinde kendinizi ne sıklıkta, gergin ve stresli hissettiniz?

4- Son bir ay içinde, yaşamınızdaki can sıkıcı durumlarla ne sıklıkta başarılı bir biçimde baş ettiniz? 

5- Son bir ay içinde ne sıklıkta, yaşamınızda meydana gelen önemli değişikliklerle etkili bir biçimde başa çıktığınızı hissettiniz?

6- Son bir ay içinde ne sıklıkta, kişisel sorunlarınızla baş etme yeteneğinizden emin oldunuz?

7- Son bir ay içinde ne sıklıkta, işlerin istediğiniz gibi gittiğini hissettiniz?

8- Son bir ay içinde ne sıklıkta, yapmak zorunda olduğunuz her şeyin üstesinden gelemeyeceğinizi düşündünüz?

9- Son bir ay içinde yaşamınızdaki rahatsız edici olayları ne sıklıkta kontrol edebildiniz?

10- Son bir ay içinde ne sıklıkta, yaşamınızdaki olaylara hakim olduğunuzu hissettiniz?

11- Son bir ay içinde, kontrolünüz dışında gerçekleşen şeylerden dolayı ne sıklıkta öfkelendiniz?

12- Son bir ay içinde ne sıklıkta, üstesinden gelmek zorunda olduğunuz şeyler üzerinde düşündünüz?

13- Zamanınızı nasıl geçirdiğinizi son bir ay içinde ne sıklıkta kontrol edebildiniz?

14- Son bir ay içinde ne sıklıkta, güçlüklerin, üstesinden gelemeyeceğiniz kadar çoğaldığını hissettiniz?\newline\newline
\textbf{Ters puanlanan maddeler:} 4, 5, 6, 7, 9, 10, 13\newline
\textbf{Faktör I: Yetersiz özyeterlik algısı:} 4, 5, 6, 8, 9, 10, 13\newline
\textbf{Faktör II: Stres/rahatsızlık algısı:} 1, 2, 3, 7, 11, 12, 14
}

\subsection{Kullanılan Skor Sistemi}
Bu çalışma, yapay zeka ödevi kapsamında bireyin stresli olup olmadığını farklı bir yöntem ile değerlendirmeyi amaçlamaktadır. Çalışmada testi tamamlayan yeni bir kişinin test sonucunun makine tarafından tahmin edilip edilemeyeceği ve cevaplandırılan soruların kapsamına göre stresin türünün sınıflara ayrılıp ayrılamayacağı değerlendirilmiştir. 

Ölçekteki puanlama sistemi algoritma olarak kurgulanarak, kapalı olarak tüm veri setine işlenmiştir. \textit{skor}, \textit{faktor\_1\_skor} ve \textit{faktor\_2\_skor} sütunları; stres puanı, faktör I puanı ve faktör II puanını temsil etmektedir. \textit{stres}, \textit{faktor\_1} ve \textit{faktor\_2} ise, bizim bir eşik değeri belirleyerek sonradan eğitim için oluşturduğumuz etiketlere denk gelmektedir. \textit{Stres} için eşik değeri 0-56'nın ortalaması olan 28 iken, \textit{faktör I} ve \textit{faktör II} için 14'tür. Bunun sebebi her bir faktörden 7'şer soru vardır ve yine ortalama puan 14'e denk gelmektedir. Bu seçimi yaparken genelgeçer bir yöntem kullanmış olsakta popülasyon üzerinde anlamlı bir dağılımın ortaya çıkmasını sağlamıştır.\newline\newline
Bu etiketleri kullanarak sistem ayrı ayrı değil çoklu etiketli (multi-label) olarak değerlendirilecektir. Dolayısıyla birey stres seviyesi eşiğin üstünde olmamasına rağmen faktör I ve/veya faktör II özelliklerini model üzerinde de gösterebilecektir.

\begin{figure}[htb!]
\begin{center}
\includegraphics[width=\linewidth]{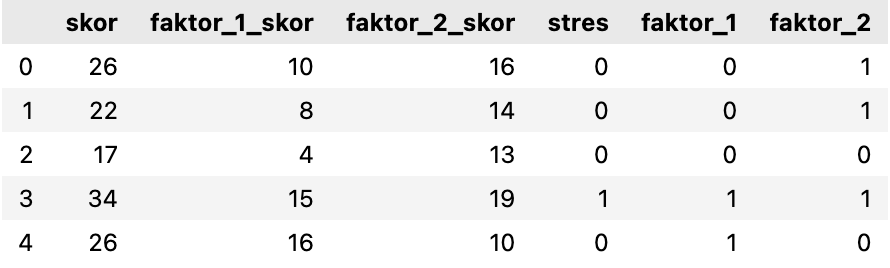}
\caption{Hesaplanan puan ve etiket yapısı.} \label{Fig:fig1}
\end{center}
\end{figure}

\section{Ver{\footnotesize İ} Anal{\footnotesize İ}z{\footnotesize İ}}

Bu makalede, stres testi sonuçları üzerinde keşifsel veri analizi gerçekleştirildi. Öncelikle, veri setinin yapısı incelendi ve popülasyonun farklı durumlar için dağılımı analiz edildi. Ardından, her bir sorunun stres düzeyini ne ölçüde etkilediğini anlamak amacıyla çeşitli analizler yapıldı (bu analizler daha ayrıntılı bir şekilde makine öğrenmesi bölümünde açıklanacaktır).

Bu analizler, popülasyon yapısını açıklamada yardımcı olabileceği gibi pek çok içgörü sağlamaktadır. Dolayısıyla çeşitli dağılımları görselleştirmek ve yorumlamak için bu grafikleri de çıkardık.

\begin{figure}[htb!]
\centering
\subfloat[\centering Genel Stres Etiket Dağılımı\label{stresLabel}]
{\includegraphics[width=0.16\textwidth]{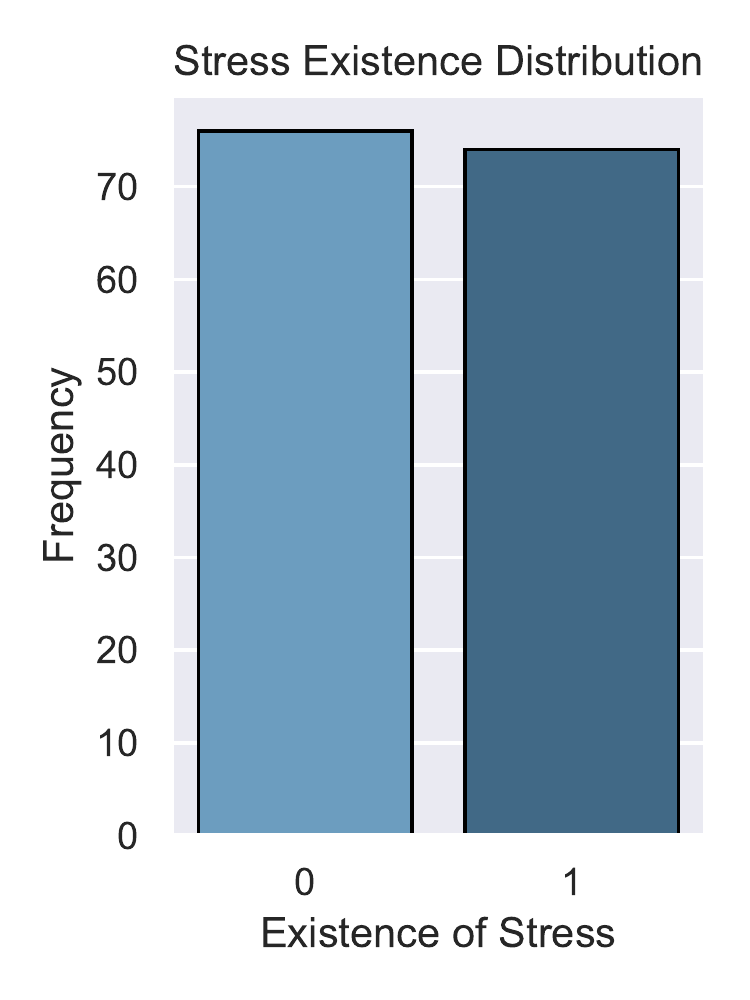} }%
    \subfloat[\centering Faktör I Etiket Dağılımı\label{f1Label}]{{\includegraphics[width=0.16\textwidth]{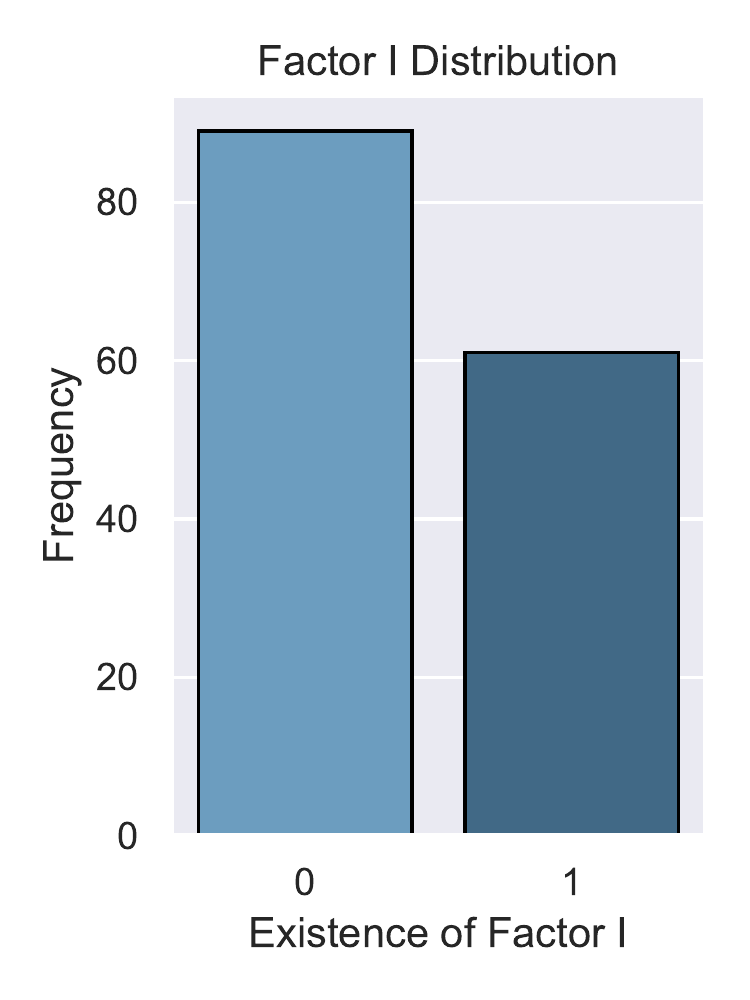} }}
    \subfloat[\centering Faktör II Etiket Dağılımı\label{f2Label}]{{\includegraphics[width=0.16\textwidth]{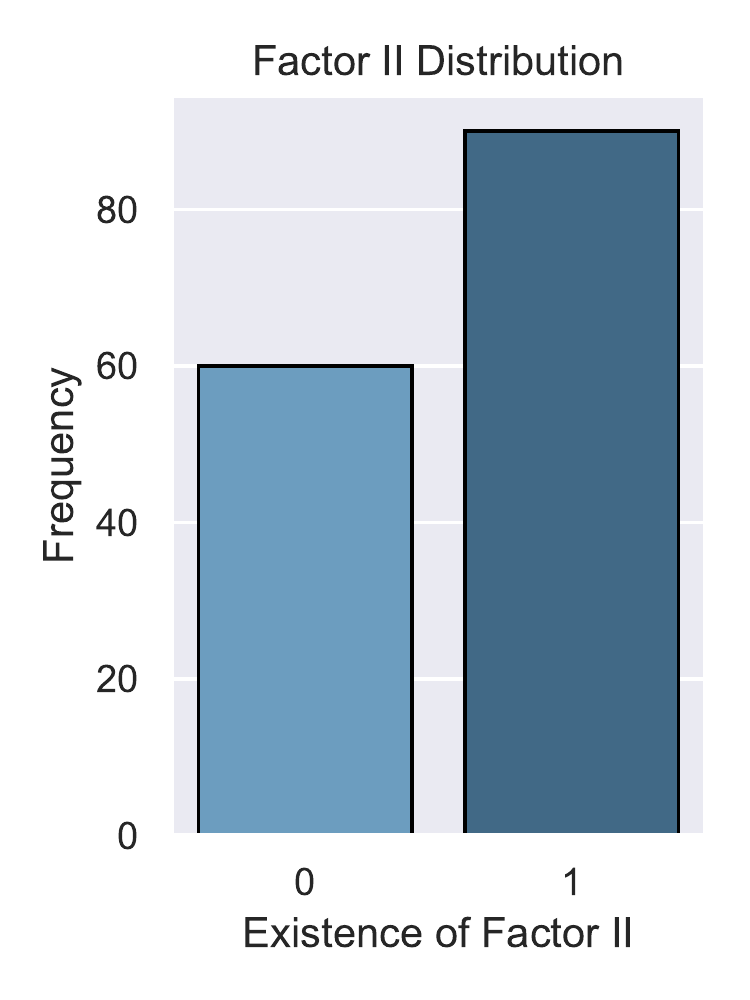} }}
\caption{Etiket dağılımları.} \label{fig:etiket}%
\end{figure}

\begin{figure}[htbp]
\centering
\subfloat[\centering Genel Stres Skor Dağılımı\label{stresScore}\vspace{0.33cm}]
{\includegraphics[width=0.49\textwidth]{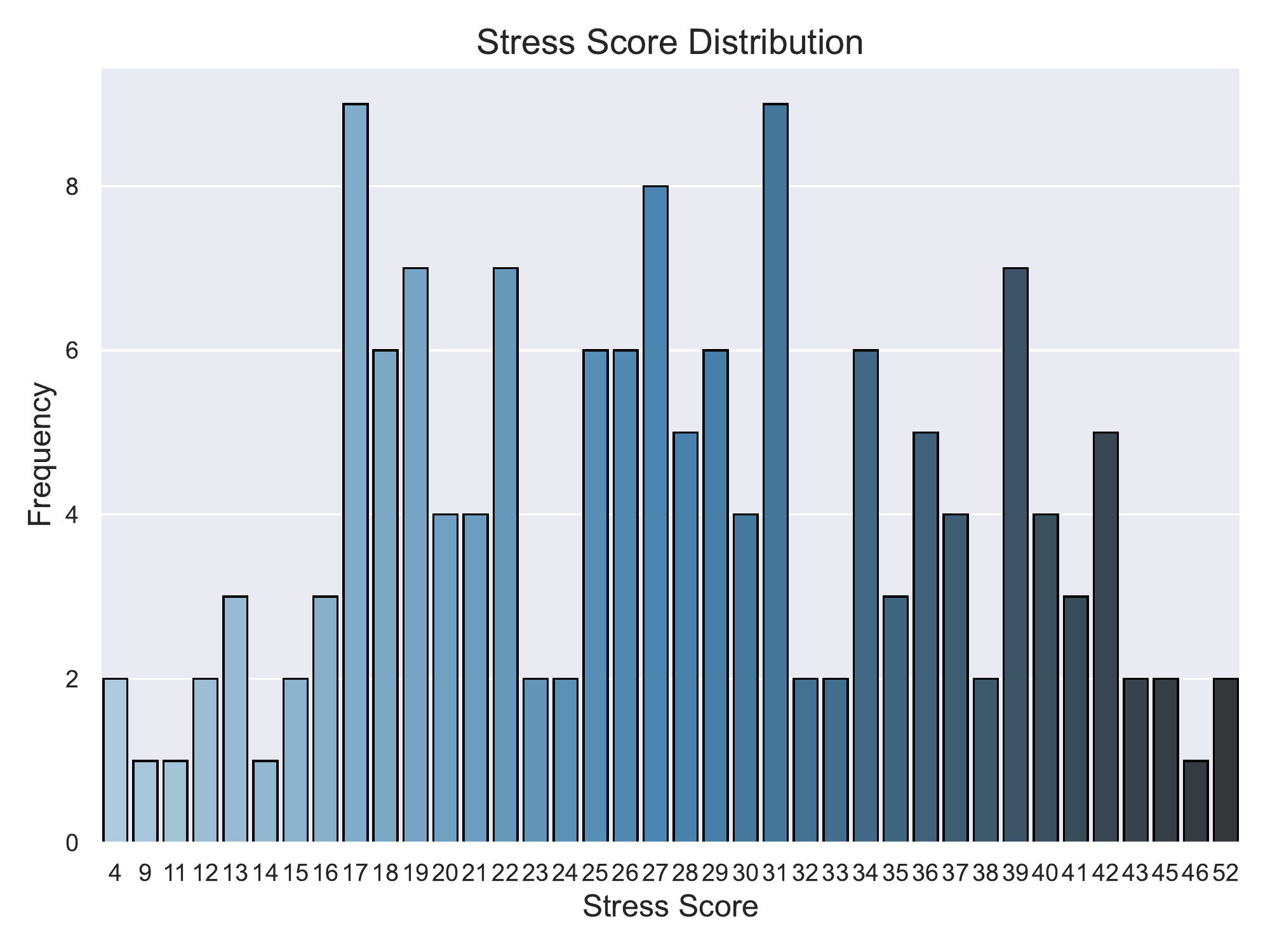} }%
    \quad
    \subfloat[\centering Faktör I Skor Dağılımı\label{f1score}\vspace{0.33cm}]{{\includegraphics[width=0.49\textwidth]{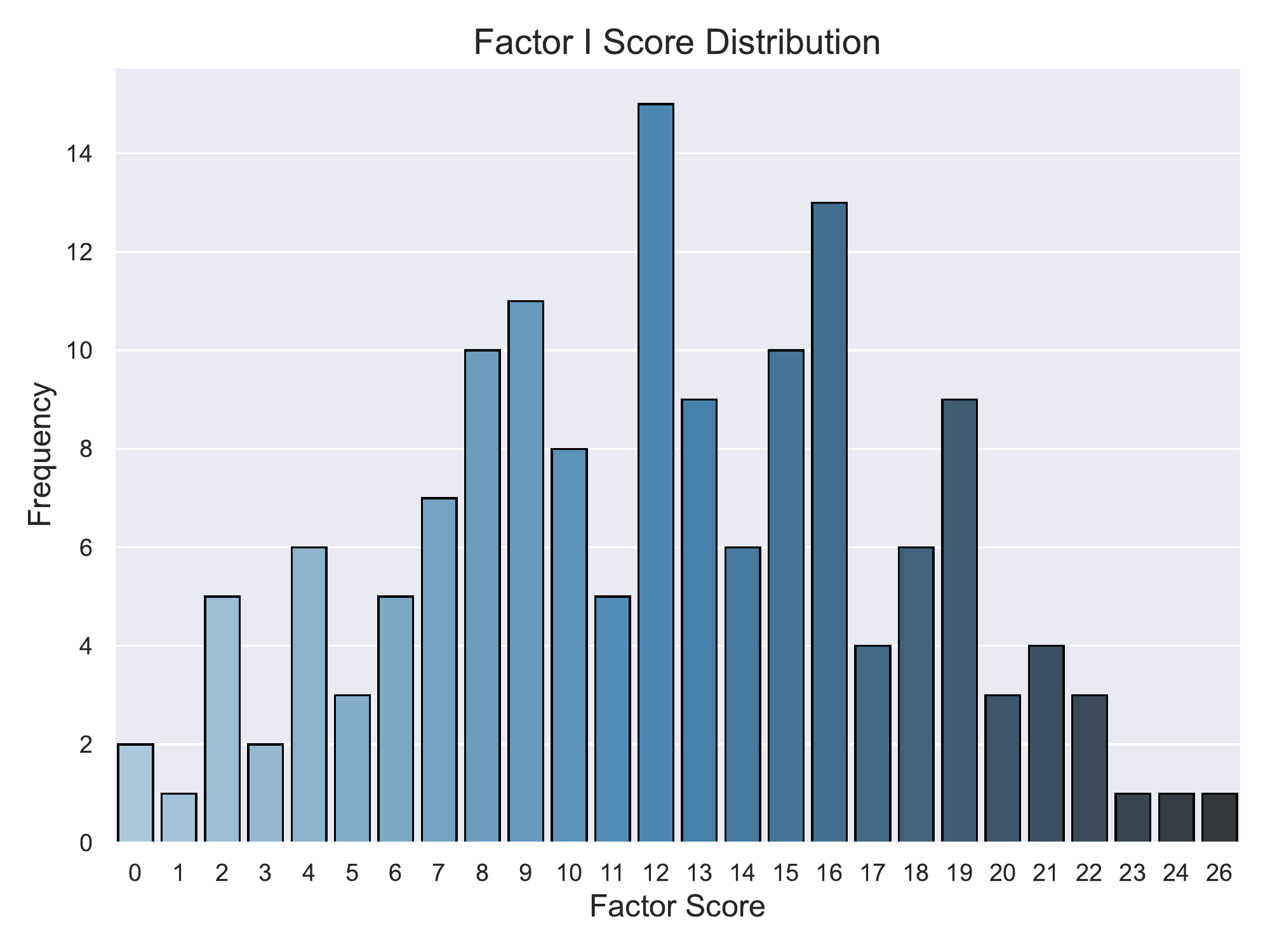} }}
    \quad
    \subfloat[\centering Faktör II Skor Dağılımı\label{f2score}]{{\includegraphics[width=0.49\textwidth]{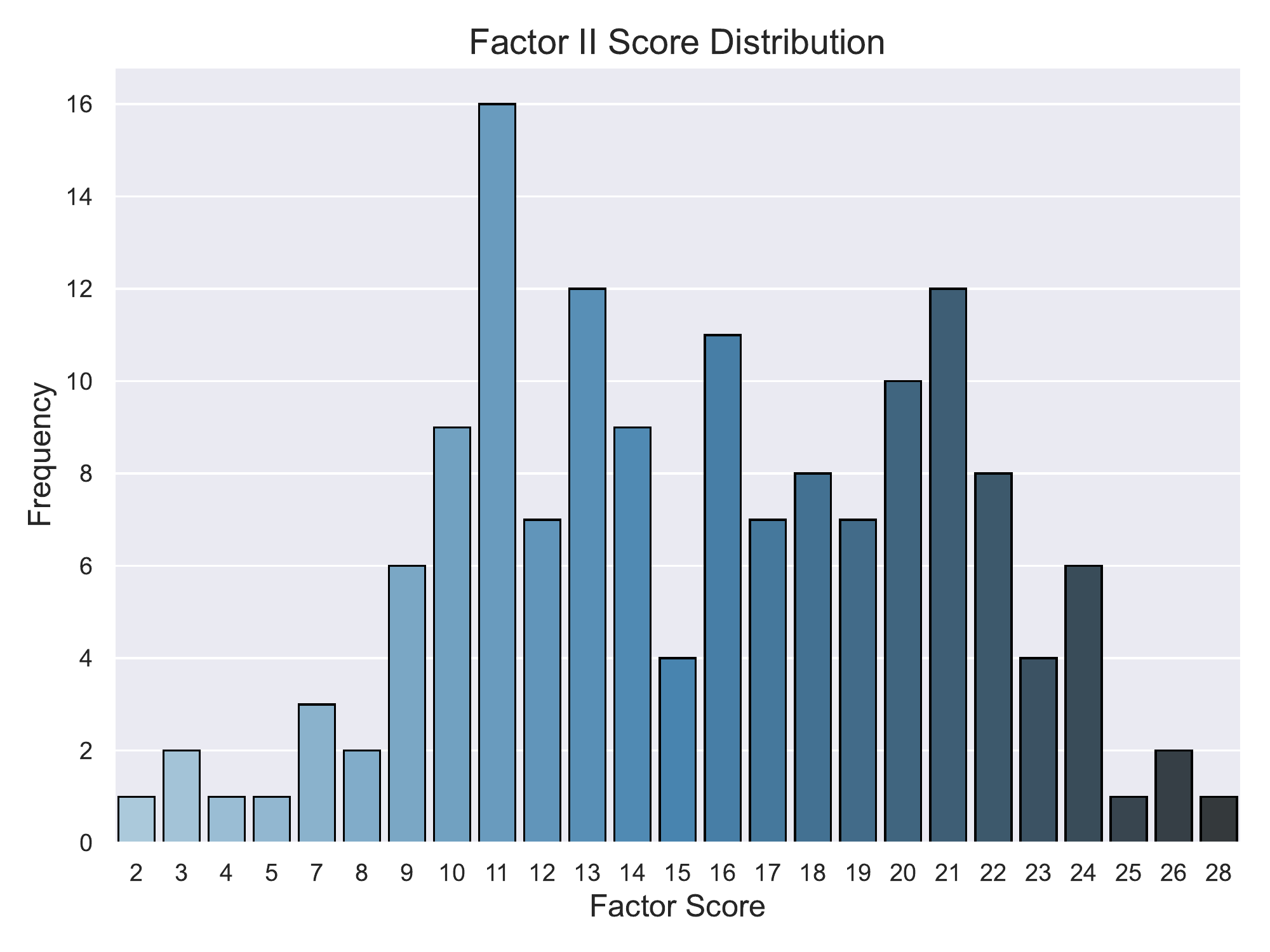} }}
\caption{Popülasyon yapısı.} \label{fig:populasyon}%
\end{figure}

\renewcommand{\arraystretch}{1.2}
\begin{table}[htb!]
\centering
\resizebox{0.36\textwidth}{!}{%
\begin{tabular}{cccc}
\hline
      & \textbf{Genel Stres} & \textbf{Faktör I} & \textbf{Faktör II} \\ \hline
count & 150                  & 150               & 150                \\
mean  & 27.72                & 12.12             & 15.60              \\
std   & 9.65                 & 5.56              & 5.35               \\
min   & 4                    & 0                 & 2                  \\
25\%  & 20                   & 8                 & 11                 \\
50\%  & 27                   & 12                & 16                 \\
75\%  & 35                   & 16                & 20                 \\
max   & 52                   & 26                & 28                
\end{tabular}}
\caption{Popülasyonun istatiksel dağılımı.}\label{istatistik}
\end{table}

\renewcommand{\arraystretch}{1.2}
\begin{table}[htb!]
\centering
\resizebox{0.28\textwidth}{!}{%
\begin{tabular}{cccc}
\hline
  & \textbf{Stres} & \textbf{Faktör I} & \textbf{Faktör II} \\ \hline
0 & 76             & 89                & 60                 \\
1 & 74             & 61                & 90                
\end{tabular}}
\caption{Popülasyonun etiket dağılımı.}
\end{table}

Faktör I ve faktör II (Fig. \ref{f1Label} ve \ref{f2Label}) etiket dağılımları görece dengesizken, genel stres (Fig. \ref{stresLabel}) popülasyonda eşit olarak dağılmıştır. Faktör I soruları ağırlıklı olarak özyeterlikle alakalıyken, Faktör II sorularının sinir/stres ile doğrudan alakalı olması bu aradaki farkı açıklamaktadır. Tablo \ref{istatistik}'e bakıldığında teste katılanların detaylı istatistiksel bilgileri görülebilmektedir.
Fig. \ref{stresScore}'ya bakıldığında, genel stres düzeyinin popülasyon üzerinde orantılı dağıldığı görülebilmektedir. Fig. \ref{f1score} ve \ref{f2score} analize dahil edildiğinde ise kişilerin farklı nedenlerle stresli değerlendirilebildiği gösterilmektedir. 

\section{Mak{\footnotesize İ}ne Öğrenmes{\footnotesize İ}}

Sunduğumuz hipotez, her bir sorunun stres düzeyini belirtmede eşit nitelikte olmadığını, makine öğrenmesi yöntemlerinin bu farkı ortaya çıkarabileceğini öne sürmektedir. Bununla birlikte, makine öğrenme modeli, hangi sorunun neyi ifade ettiği hakkında bilgi sahibi olmadığından, yalnızca tarafımızdan belirlenen 0 ve 1 etiketlerine dayanarak kendini optimize etmektedir.

Sonuçları elde etmek için bir pipeline tasarlanmıştır ve bu tasarım, tek bir çalıştırma ile 5 farklı rastgele durum için tüm sonuçları kaydedebilmektedir. Bu pipeline, XGBoost, Random Forest, Decision Tree, Gradient Boosting, CatBoost ve AdaBoost olmak üzere 6 farklı makine öğrenmesi modelinden oluşmaktadır. Her bir model, veri setinin 5 farklı rastgele durumu için sonuçları elde edilmiş ve raporlanmıştır. Modeller, sklearn'in \verb|MultiOutputClassifier| sınıfı kullanılarak çok-etiketli yapıda çalışacak şekilde düzenlenmiştir. Sonuçların değerlendirilmesinde, makro kesinlik (macro precision), makro geri çağırma (macro recall) ve makro F1 skoru tercih edilmiştir. Makro yaklaşımının tercih edilme sebebi, faktör I ve faktör II etiketlerinin dengeli bir dağılıma sahip olmamasıdır. Bu durumu göz ardı etmemek ve her etiketin sonuç üzerinde eşit etkiye sahip olmasını sağlamak için "makro" yaklaşımı kullanılmaktadır.

\begin{equation}
    \small\texttt{Macro Precision} = \frac{\sum_{n} \big(\frac{TP}{TP+FP}\big)}{\small\texttt{sınıf sayısı}\:(n)}
    \label{eq:macropre}
\end{equation} 
\begin{equation}
    \small\texttt{Macro Recall} =  \frac{\sum_{n} \big(\frac{TP}{TP+FN}\big)}{\small\texttt{sınıf sayısı}\:(n)} 
\label{eq:macrorec}
\end{equation} 
\begin{equation}
    \small\texttt{Macro F1 Score} = \frac{\sum_{n} \big(\frac{2*TP}{2*TP+FP+FN}\big)}{\small\texttt{sınıf sayısı}\:(n)}
\label{eq:macrof1}
\end{equation}

K-fold ve benzeri yaklaşımların bu çalışma için veri sayısının yeterli olmaması gibi sebeplerle uygun olmadığını düşündüğümüz için daha önce farklı bir alanda kullandığımız yaklaşım ile cross-validation gerçeklenmiştir \cite{tanyel4421493deciphering}. Aynı zamanda özellikler incelenirken, PCA gibi yöntemler ile istatistiksel veri kaybına uğramaya gerek kalmadan çözülebilecek bir problem ile çalıştığımız için PCA ve benzeri yöntemler de kullanılmamıştır. Ancak özellik analizi ve test kapsamında sorulan soruların modeller üzerindeki etkisi detaylı ve makine öğrenmesi literatürüne uygun şekilde incelendi.

\renewcommand{\arraystretch}{1.2}
\begin{table}[htb!]
\centering
\resizebox{0.45\textwidth}{!}{%
\begin{tabular}{cccc}
\textbf{Model} & \textbf{Precision}    & \textbf{Recall}       & \textbf{F1 Skoru}    \\ \hline
\textbf{ADA}    & \textbf{92.53 ± 5.07} & 91.96 ± 3.13          & \textbf{92.05 ± 2.93} \\
\textbf{CB}     & 90.67 ± 2.70          & 91.07 ± 5.69          & 90.71 ± 4.04          \\
\textbf{DT}     & 86.86 ± 4.08          & 90.56 ± 2.02          & 88.29 ± 1.55          \\
\textbf{GB}     & 89.54 ± 3.25          & 90.59 ± 5.09          & 89.88 ± 3.54          \\
\textbf{RF}     & 91.71 ± 2.33          & 90.85 ± 4.27          & 91.17 ± 3.09          \\
\textbf{XGB}    & 90.68 ± 4.53          & \textbf{92.81 ± 5.12} & 91.58 ± 3.93         
\end{tabular}}
\caption{Veri setinin 5 farklı rastgele durumu için 6 farklı modelin ortalama multilabel sonuçları.}\label{perf}
\end{table}

\subsection*{Makine öğrenmesi sonuçlarının değerlendirilmesi}

\begin{figure*}[htbp]
\centering
\subfloat[\centering Durum 1]
{\includegraphics[width=0.61\textwidth]{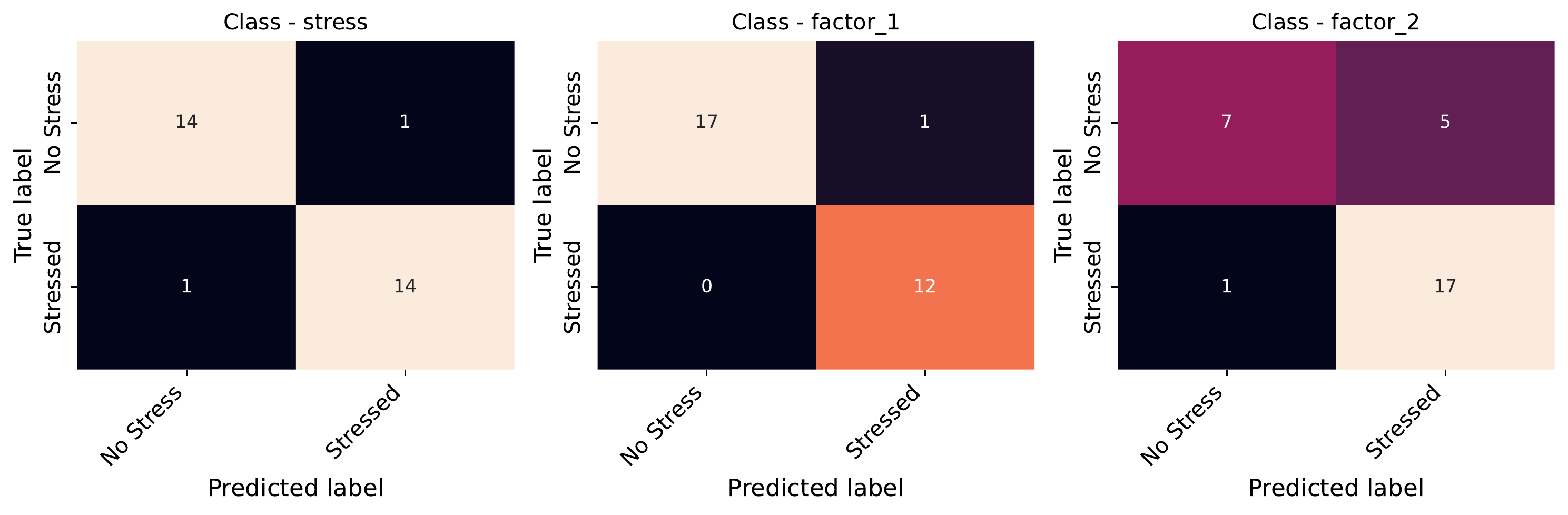} }%
    \quad
    \subfloat[\centering Durum 2]{{\includegraphics[width=0.61\textwidth]{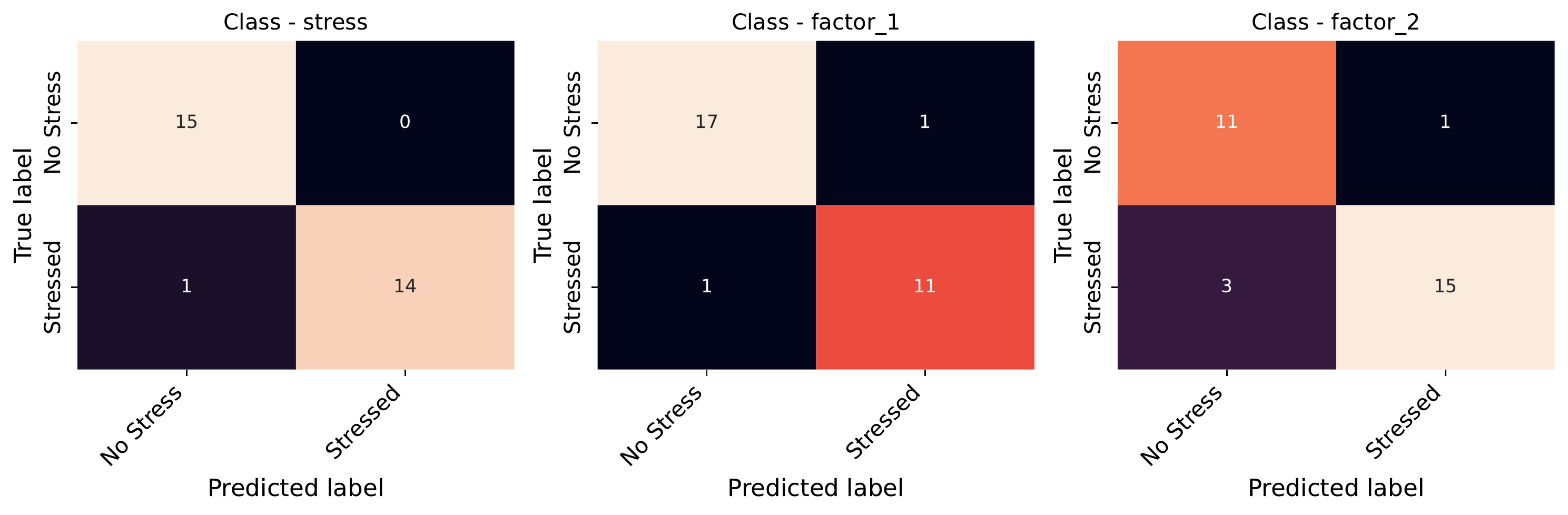} }}
    \quad
    \subfloat[\centering Durum 3]{{\includegraphics[width=0.61\textwidth]{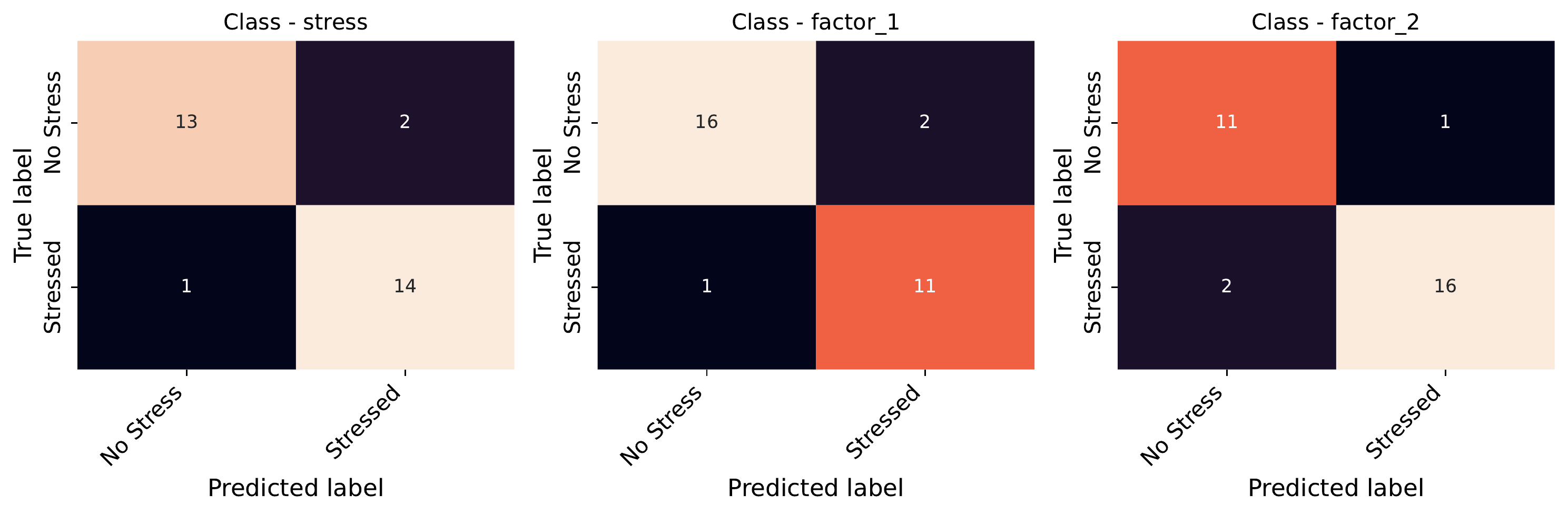} }}
    \quad
    \subfloat[\centering Durum 4]{{\includegraphics[width=0.61\textwidth]{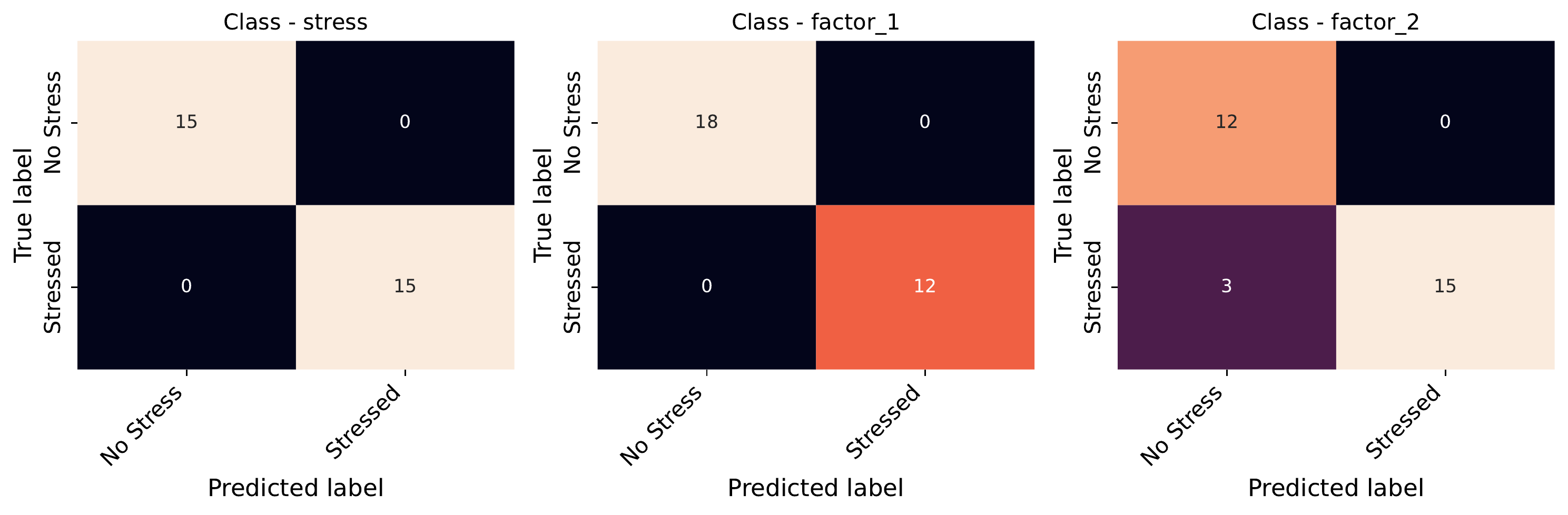} }}
    \quad
    \subfloat[\centering Durum 5]{{\includegraphics[width=0.61\textwidth]{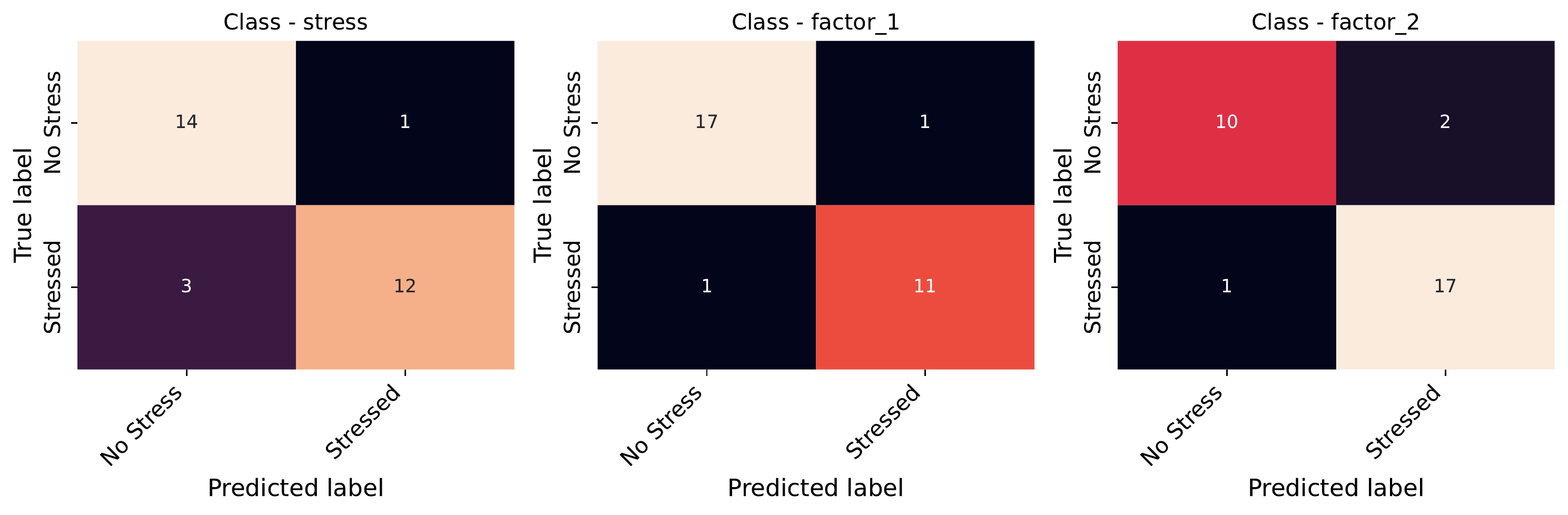} }}
\caption{AdaBoost, 5 farklı dağılım için karmaşıklık matrisleri.} \label{fig:adaConf}%
\end{figure*}

\begin{figure*}[htbp]
\centering
\subfloat[\centering Durum 1]
{\includegraphics[width=0.61\textwidth]{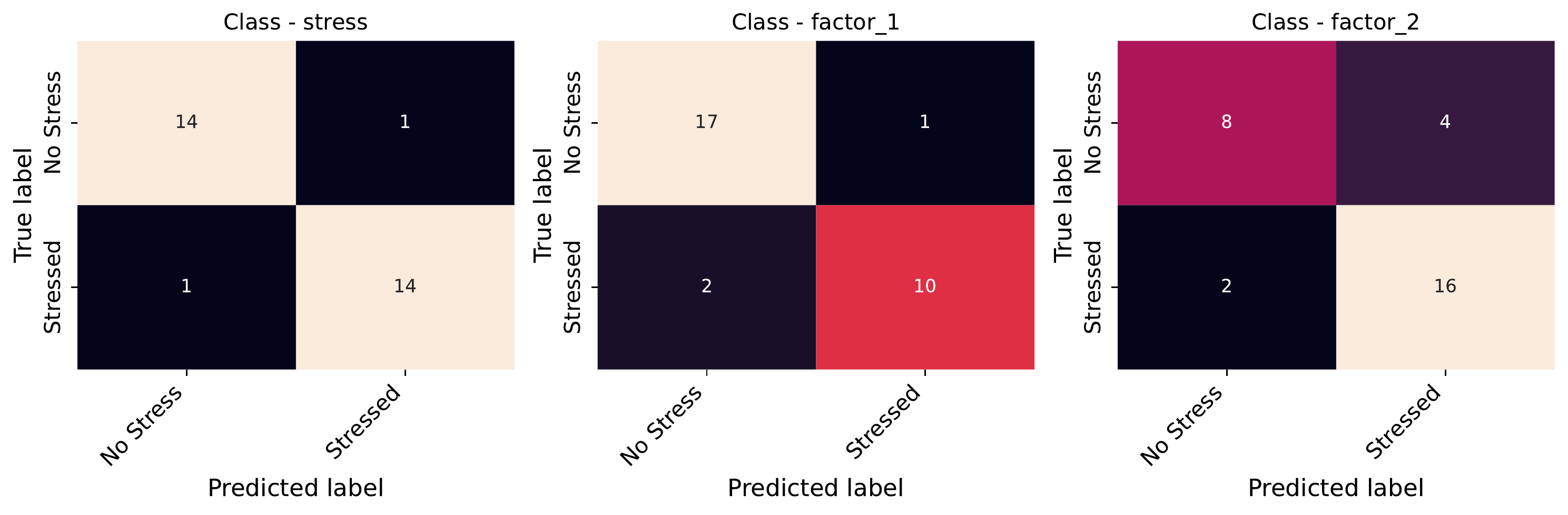} }%
    \quad
    \subfloat[\centering Durum 2]{{\includegraphics[width=0.61\textwidth]{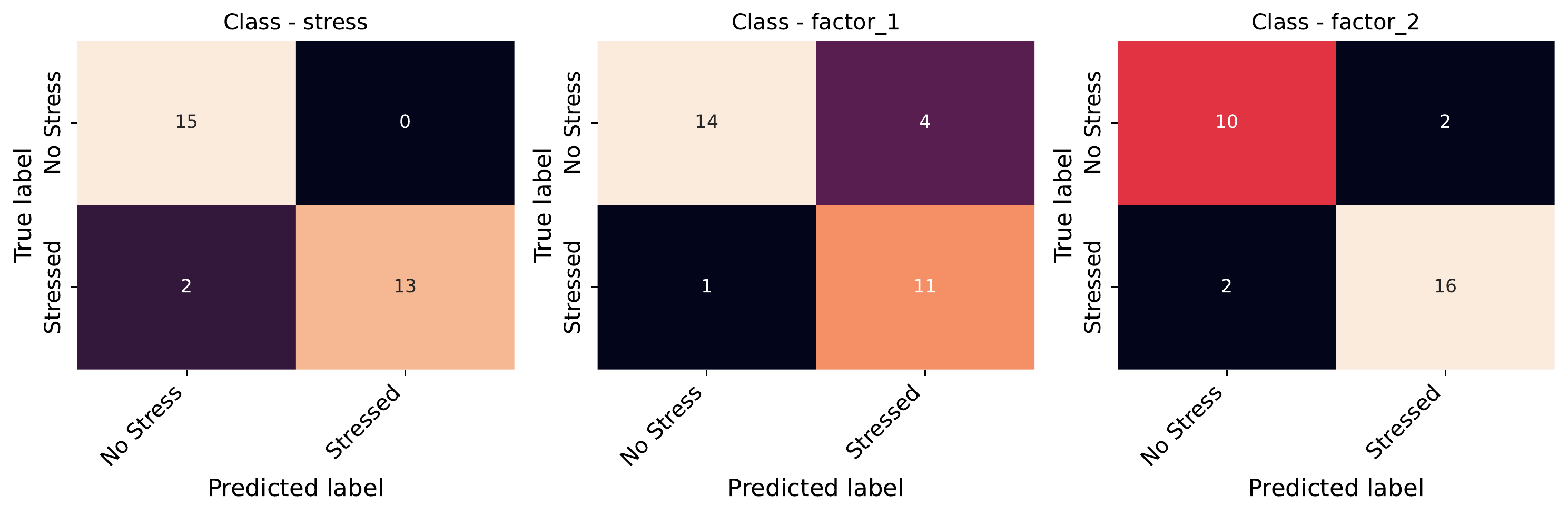} }}
    \quad
    \subfloat[\centering Durum 3]{{\includegraphics[width=0.61\textwidth]{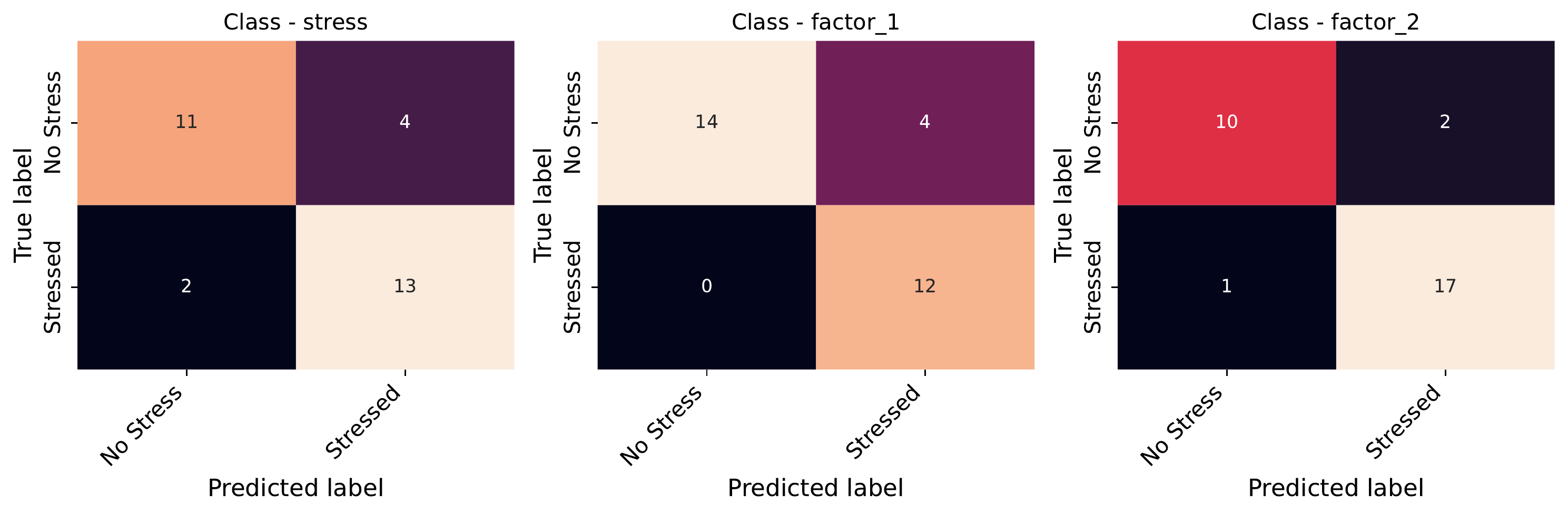} }}
    \quad
    \subfloat[\centering Durum 4]{{\includegraphics[width=0.61\textwidth]{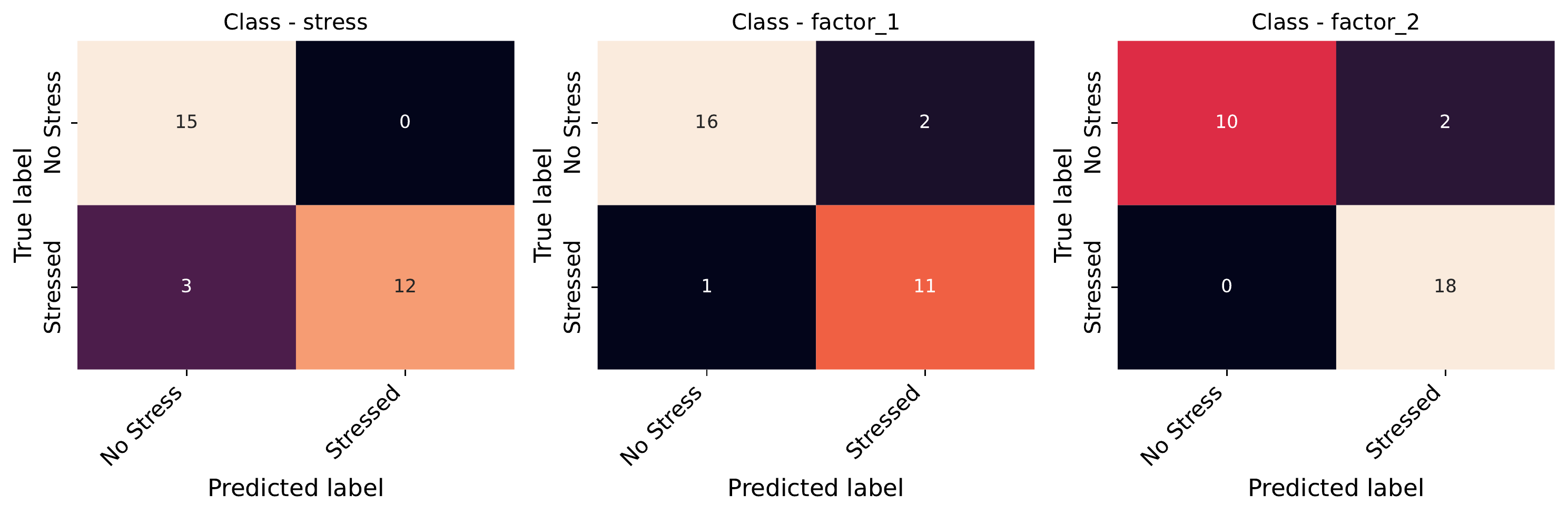} }}
    \quad
    \subfloat[\centering Durum 5]{{\includegraphics[width=0.61\textwidth]{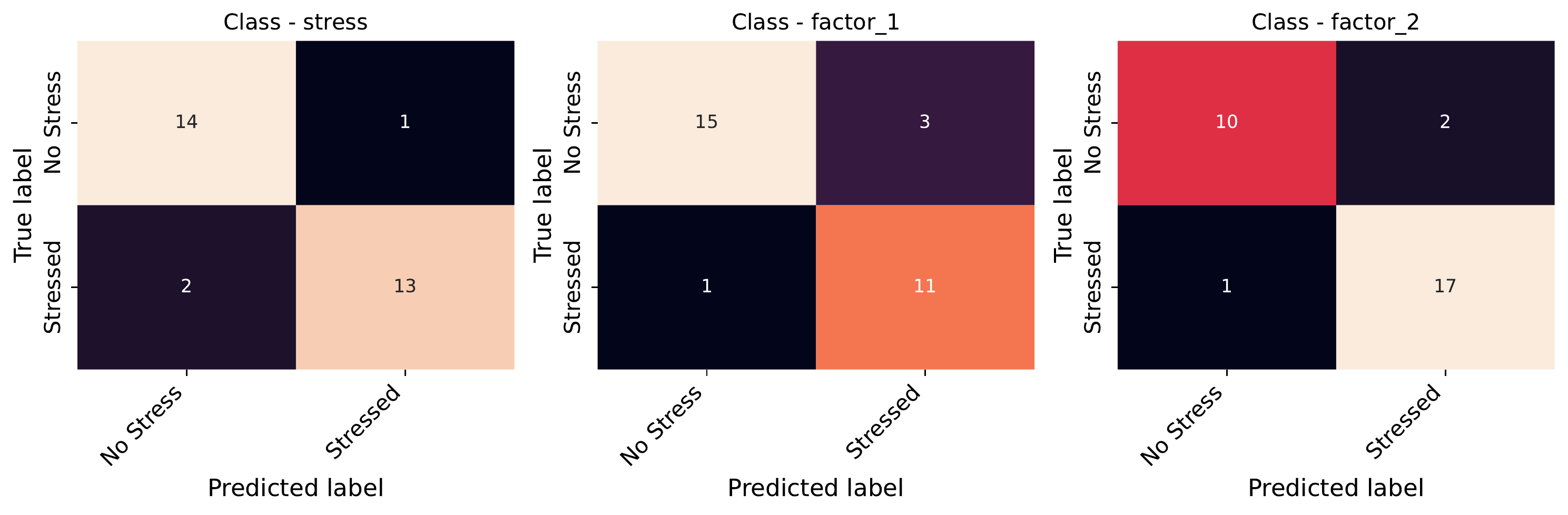} }}
\caption{Decision Tree, 5 farklı dağılım için karmaşıklık matrisleri.} \label{fig:dtconf}%
\end{figure*}

En başarılı olan model AdaBoost iken en başarısız olan model Decision Tree olarak karşımıza çıkmaktadır. Ayrıca bu modeller için "karmaşıklık matrisi" (Fig. \ref{fig:adaConf} ve \ref{fig:dtconf}) çıkarılarak daha detaylı olarak incelenmiştir. Buna ek olarak, CatBoost, Random Forest ve XGBoost modelleri de \%90 başarımı geçmektedir. Bu durum, makine öğrenmesi modellerinin hiçbir kural vermesek dahi elimizdeki popülasyonun sorulara verdiği yanıt ile stres durumu arasındaki ilişkiyi kullanılan mevcut test üzerinden yakalayabildiğini göstermektedir.

Farklı durumlarda aynı modelin karmaşıklık matrisleri incelendiğinde (Fig. \ref{fig:adaConf} ve \ref{fig:dtconf}), test setine giden bireyler değiştiğinde tahmin başarısındaki değişikliğin nasıl etkilendiği görülmektedir. 

Stres durumu AdaBoost modelinde ilk durumda 2, ikinci durumda 1, üçüncü durumda 3, dördüncü durumda 0 ve beşinci durumda 4 kişi için hata yapmıştır. Faktör I ve Faktör II tahmini daha zor gibi gözükse de Faktör I'in neredeyse tüm durumlarda stres durumundan daha doğru tahmin edildiği görülebilmektedir. Model Faktör II'de ise birinci ve ikinci durumda 6 ve 4 hata, kalan durumlarda ise 3 kişide hata yapmaktadır.

Stres durumu Decision Tree modelinde ilk durumda 2, ikinci durumda 2, üçüncü durumda 6, dördüncü durumda 3 ve beşinci durumda 3 kişi için hata yapmıştır. Bu model için Faktör I tespiti de AdaBoost'a göre daha zayıftır. Faktör I için yaptığı hata sayıları, 3, 5, 4, 3 ve 4 iken Faktör II'de 6, 4, 3, 2 ve 3'tür.

\begin{figure*}[htbp]
\centering
\subfloat[\centering\label{xgb}]
{\includegraphics[width=0.45\textwidth]{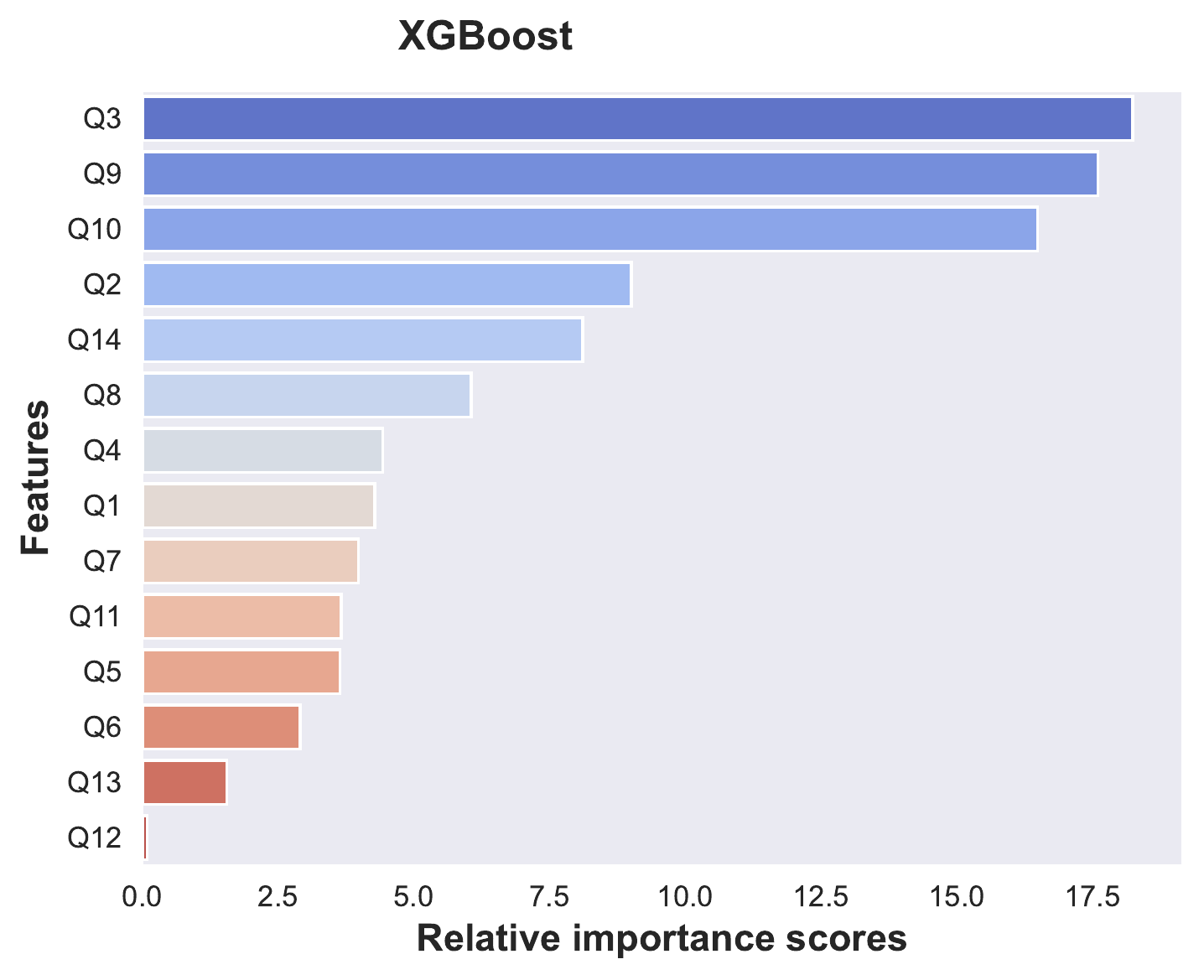} }%
    \subfloat[\centering\label{rf}]{{\includegraphics[width=0.45\textwidth]{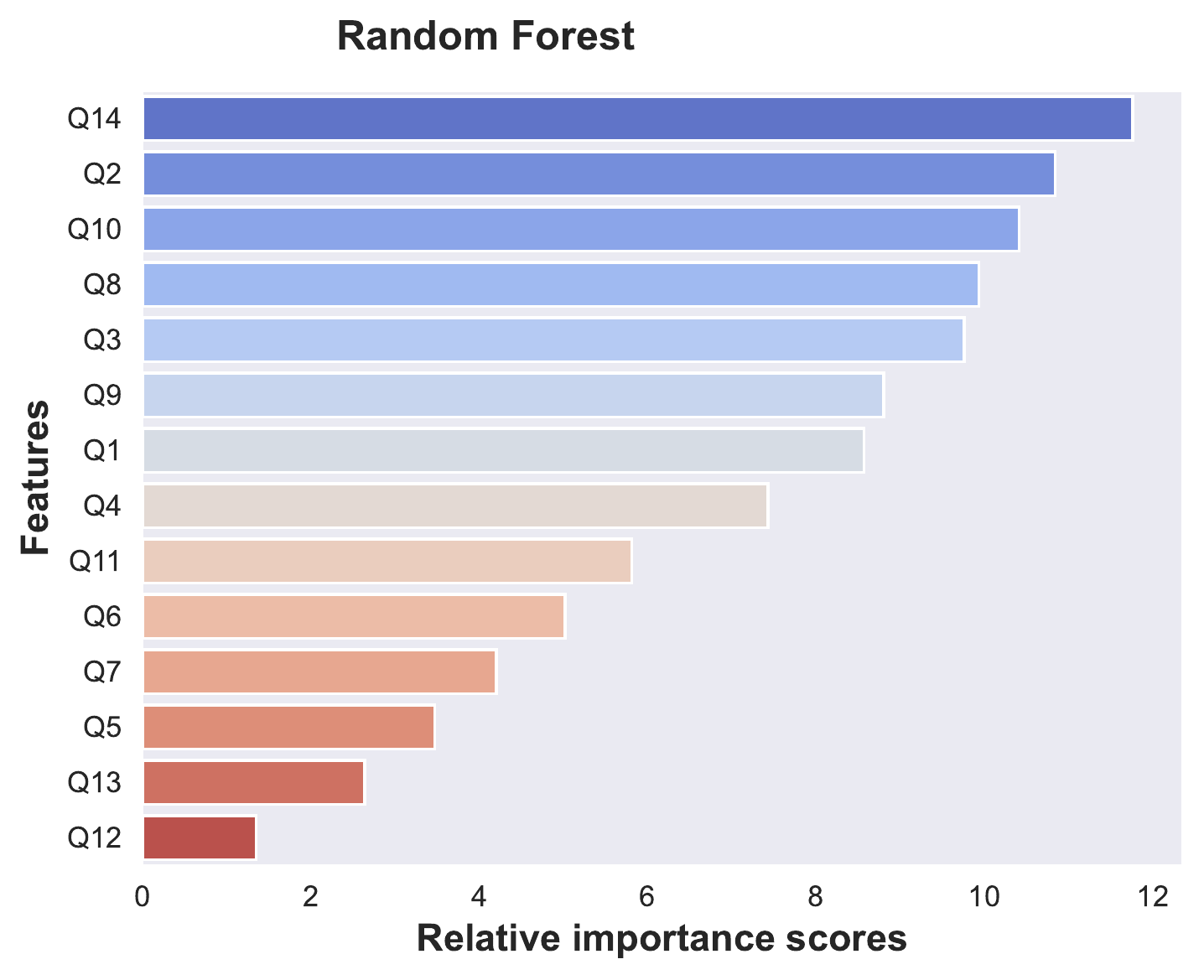} }}
    \quad
    \subfloat[\centering\label{dt}]{{\includegraphics[width=0.45\textwidth]{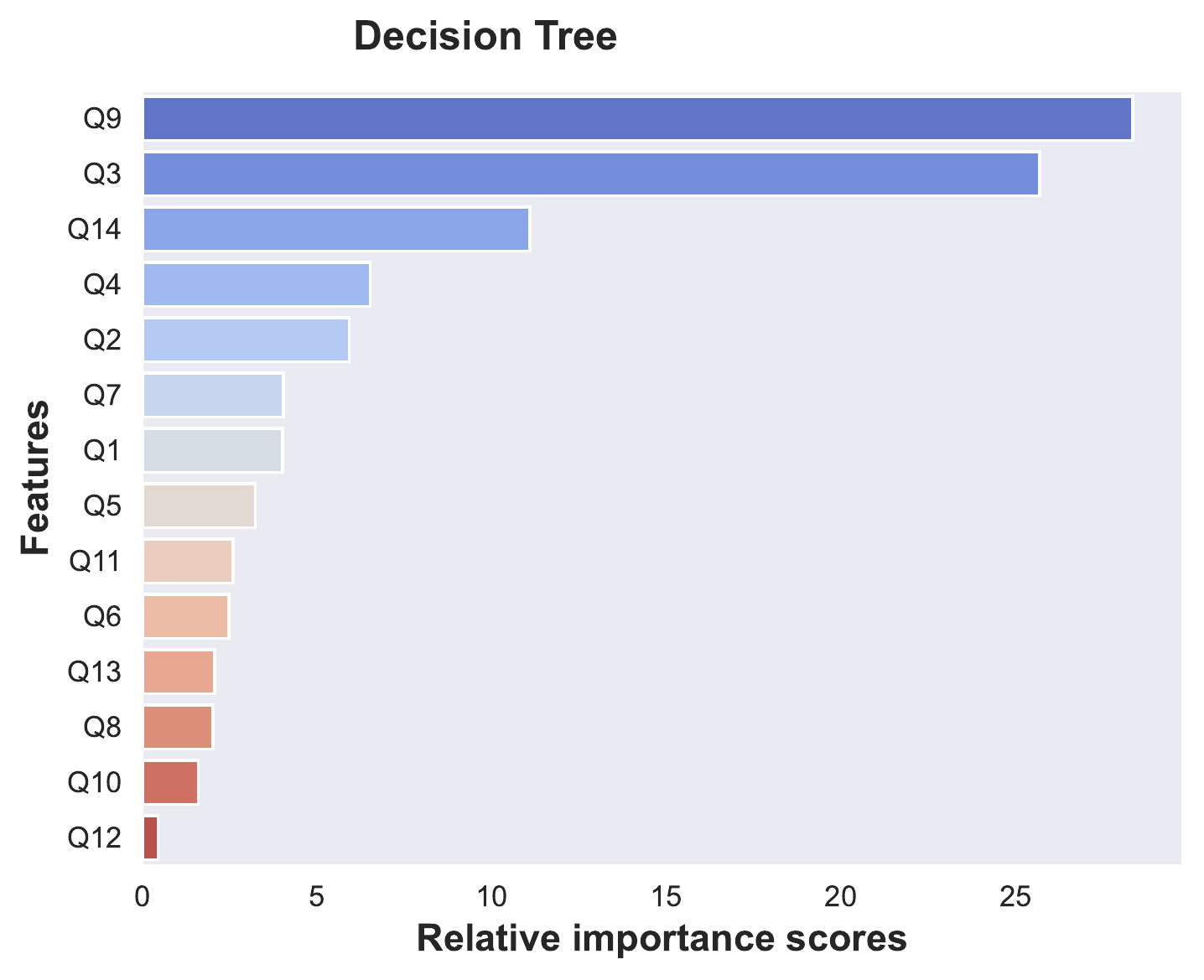} }}
    \subfloat[\centering\label{gb}]{{\includegraphics[width=0.45\textwidth]{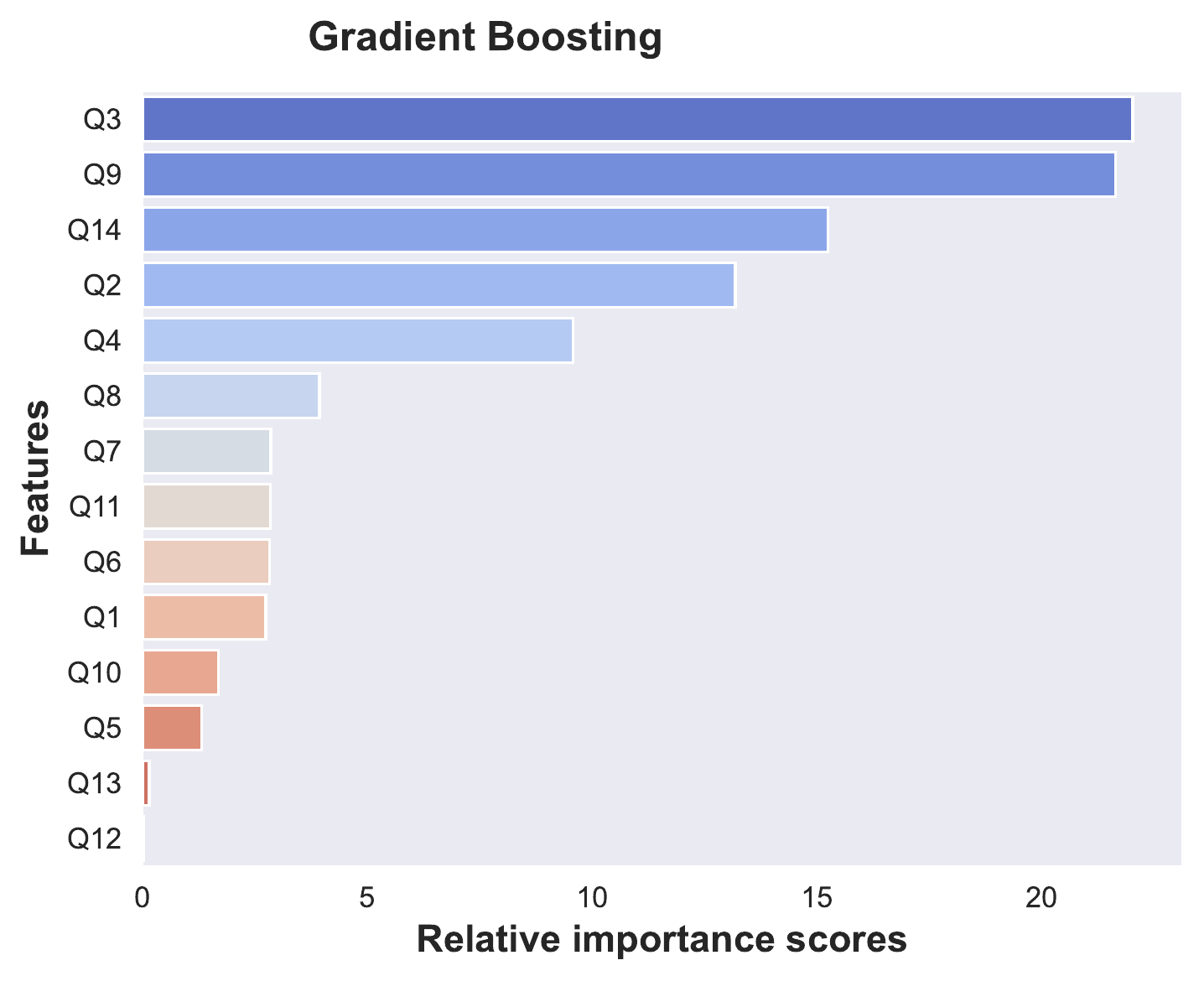} }}
    \quad
    \subfloat[\centering\label{cb}]{{\includegraphics[width=0.45\textwidth]{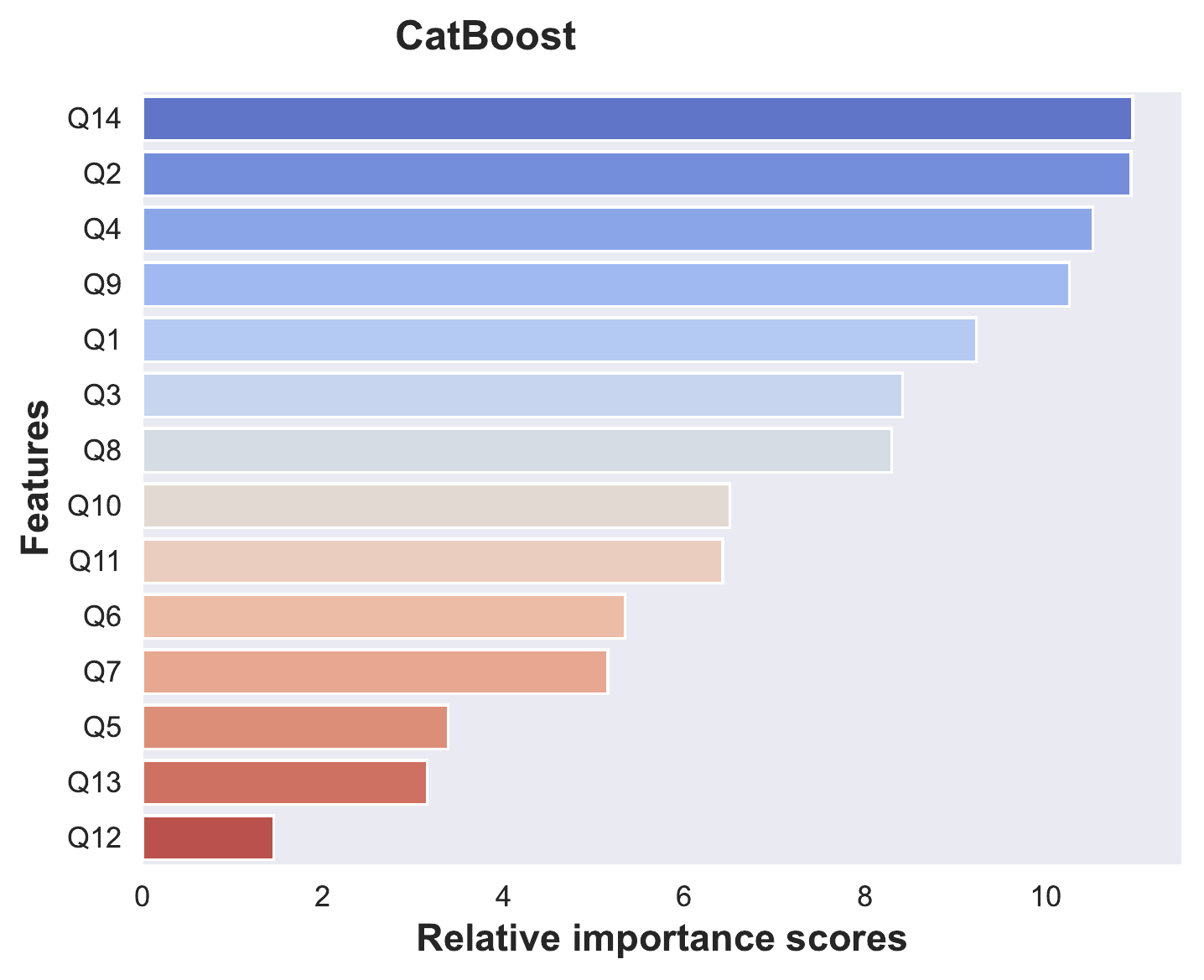} }}
    \subfloat[\centering\label{ada}]{{\includegraphics[width=0.45\textwidth]{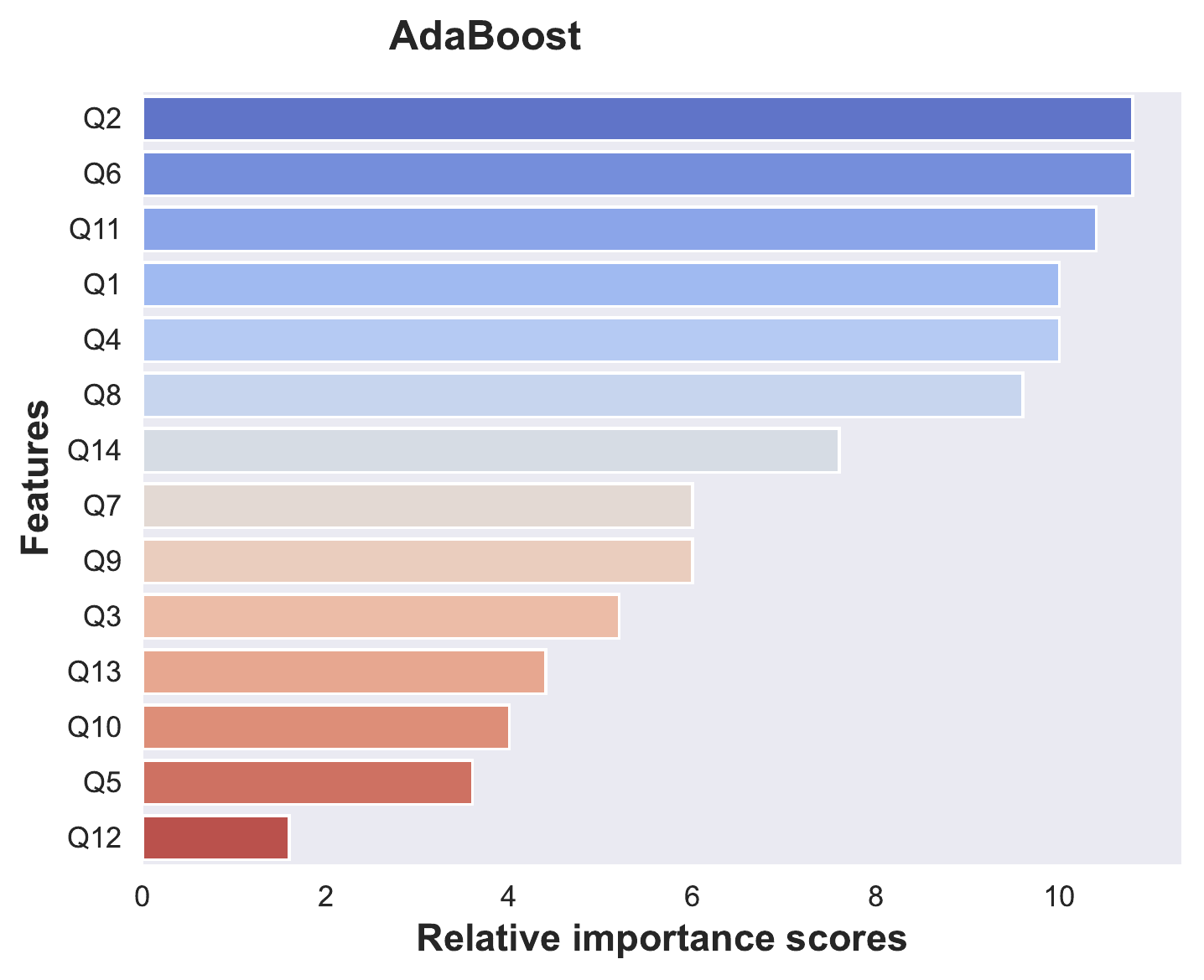} }}
\caption{Veri setinin 5 farklı rastgele durumu için 6 farklı modelin sorulara verdiği ortalama önem sonuçları.} \label{fig:ozellikAnalizi}%
\end{figure*}

\todo[inline, color=gray!5]{2- Son bir ay içinde ne sıklıkta, yaşamınızdaki önemli şeyleri kontrol edemediğinizi hissettiniz? 

\textnormal{\textcolor{darkspringgreen}{(\textit{Faktör II, Negatif Soru}, önemli)}}\newline\newline6- Son bir ay içinde ne sıklıkta, kişisel sorunlarınızla baş etme yeteneğinizden emin oldunuz? 

\textnormal{\textcolor{darkspringgreen}{(\textit{Faktör I, Pozitif Soru}, önemli)}}\newline\newline9- Son bir ay içinde yaşamınızdaki rahatsız edici olayları ne sıklıkta kontrol edebildiniz? 

\textnormal{\textcolor{darkspringgreen}{(\textit{Faktör I, Pozitif Soru}, önemli)}}\newline\newline14- Son bir ay içinde ne sıklıkta, güçlüklerin, üstesinden gelemeyeceğiniz kadar çoğaldığını hissettiniz? 

\textnormal{\textcolor{darkspringgreen}{(\textit{Faktör II, Negatif Soru}, önemli)}}\newline\newline5- Son bir ay içinde ne sıklıkta, yaşamınızda meydana gelen önemli değişikliklerle etkili bir biçimde başa çıktığınızı hissettiniz? 

\textnormal{\textcolor{red}{(\textit{Faktör I, Pozitif Soru}, az önemli/önemsiz)}}\newline

10- Son bir ay içinde ne sıklıkta, yaşamınızdaki olaylara hakim olduğunuzu hissettiniz?\newline
\textnormal{\textcolor{red}{(\textit{Faktör I, Pozitif Soru}, az önemli/önemsiz)}}\newline\newline12- Son bir ay içinde ne sıklıkta, üstesinden gelmek zorunda olduğunuz şeyler üzerinde düşündünüz?

\textnormal{\textcolor{red}{(\textit{Faktör II, Negatif Soru}, az önemli/önemsiz)}}\newline\newline13- Zamanınızı nasıl geçirdiğinizi son bir ay içinde ne sıklıkta kontrol edebildiniz?\newline
\textnormal{\textcolor{red}{(\textit{Faktör I, Pozitif Soru}, az önemli/önemsiz)}}}

\subsection{Soruların Analizi}

6 farklı modelin 5 farklı veri seti dağılımı için oluşturulmuş ortalama soru önem miktarları Fig. \ref{fig:ozellikAnalizi}'teki gibidir.\newline\newline
Fig. \ref{fig:ozellikAnalizi} incelendiğinde, Q3'ün 2 model tarafından (Fig. \ref{xgb} ve \ref{gb}), Q14'ün 2 model tarafından (Fig. \ref{rf} ve \ref{cb}), Q9 ve Q3'ün en kötü model tarafından (Fig. \ref{dt}) ve Q2 ve Q6'nın en iyi model tarafından (Fig. \ref{ada}) seçildiği görülmektedir. Q12 tüm modeller için en anlamsız soru olarak seçilmiştir ve arkasından çoğunlukla Q13 gelmektedir. Bunların yanı sıra Q10 ve Q5'de zaman zaman en az dikkate alınan sorular olarak yerlerini almışlardır.\newline\newline
Bu sonuçlar popülasyon üzerine her sorunun aynı etkiyi yaratmadığını ve belirli soruların da oldukça ayırt edici olduğunu göstermektedir. Burada yapılabilecek çıkarım, bazı soruların yapısından dolayı varyansa çok açık olmaması ve bu soruların popülasyondaki bireyleri ayırt etme konusunda anlamlı olmayabileceğidir. Modeller sorular arasındaki bu ilişkileri minimum bilgiyle keşfetmektedir. Gelecekte, makineye daha fazla bilgi sağlayarak ve daha büyük bir popülasyon kullanarak bu çalışmanın replikasyonunun yapılması, soruların insanların stresini ölçmede ne kadar etkili olduğunu tanımlamada alternatif bir göz olarak yerini alabilir.

\section{Sonuç}

Q2, Q6, Q9 ve Q14 modeller için en önemli sorular, Q5, Q10, Q12 ve Q13 en az dikkate alınan sorular olarak değerlendirilmiştir. 6 farklı model için \%(88.29-92.05) ortalama skor elde edilmiştir. Bu başarım, veri sayısı ve problemin zorluğu göz önüne alındığında oldukça önemlidir.

Bu çalışmada limitasyon olarak dikkat edilmesi gereken veri sayısının genelleştirilmiş bir sonuç için yeterli olmayabileceğidir. Bunun yanı sıra anket sırasında demografik bilgi toplanmaması sebebiyle bir gruba özel değil, anketin eriştiği her yaş ve kitleden veri toplanmıştır.

\bibliographystyle{plain}
\bibliography{references}

\end{document}